\documentclass[10pt,twocolumn,letterpaper]{article}

\usepackage{cvpr}
\usepackage{times}
\usepackage{epsfig}
\usepackage{graphicx}
\usepackage{amsmath}
\usepackage{amssymb}

\usepackage{verbatim}

\usepackage[pagebackref=true,breaklinks=true,letterpaper=true,colorlinks,bookmarks=false]{hyperref}

\cvprfinalcopy 

\ifcvprfinal\fi
\begin{document}

\title{How Useful is Self-Supervised Pretraining for Visual Tasks?}

\author{Alejandro Newell \quad Jia Deng\\
Princeton University \\
{\tt\small \{anewell,jiadeng\}@cs.princeton.edu}
}

\maketitle

\begin{abstract}
Recent advances have spurred incredible progress in self-supervised pretraining for vision. We investigate what factors may play a role in the utility of these pretraining methods for practitioners. To do this, we evaluate various self-supervised algorithms across a comprehensive array of synthetic datasets and downstream tasks.  We prepare a suite of synthetic data that enables an endless supply of annotated images as well as full control over dataset difficulty. Our experiments offer insights into how the utility of self-supervision changes as the number of available labels grows as well as how the utility changes as a function of the downstream task and the properties of the training data. We also find that linear evaluation does not correlate with finetuning performance. Code and data is available at \href{https://www.github.com/princeton-vl/selfstudy}{github.com/princeton-vl/selfstudy}.
\end{abstract}


\section{Introduction}

Self-supervised learning has the potential to revolutionize computer vision. It aims to learn good representations from unlabeled visual data, reducing or even eliminating the need for costly collection of manual labels. In the context of deep networks, the most common use of self-supervision is in pretraining a network with unlabeled data for later finetuning on a downstream task. The better the self-supervision, the better the downstream performance.

Progress on self-supervised pretraining has accelerated in recent years. In particular, self-supervised models now produce features that are comparable to or outperform those produced by ImageNet pretraining \cite{bachman2019learning,tian2019contrastive,henaff2019data,he2019momentum}. While it is currently not common to leverage such methods, wider adoption might be seen in the wake of these advances.

In this work we investigate what barriers might exist between the latest progress in self-supervision and its broader use in the field, as well as how to approach evaluation in a way that is informative and useful to practitioners.

\begin{figure}[t]
\begin{center}
   \includegraphics[width=\linewidth]{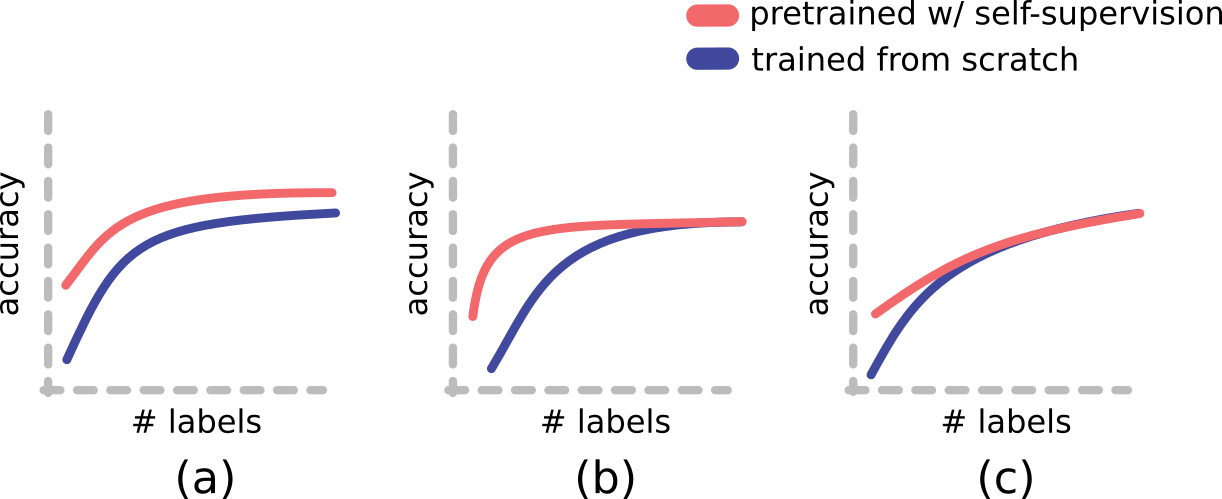}
\caption{We highlight three possible outcomes when using self-supervised pretraining, the pretrained model either: a) always provides an improvement over the the model trained from scratch even as the amount of labeled data increases, b) reaches higher accuracy with fewer labels but plateaus to the same accuracy as the baseline, c) converges to baseline performance before accuracy plateaus. In our experiments we find option (c) to be the most common outcome.}
\end{center}
\label{fig:intro_fig}
\vspace{-.7cm}
\end{figure}

We are motivated by the observation that much of the existing literature evaluates self-supervision in either few-shot settings or when restricting downstream use of the model. For example, a common form of evaluation is to freeze all weights of the pretrained network and train a linear layer for the downstream task. However, for many computer vision tasks, there already exists a large amount of labeled data, and finetuning of the model is necessary to get as high accuracy as possible.

Evaluating self-supervision in unrestricted settings is important because in practice there are many situations where maximizing accuracy is paramount (think of pedestrian detection for self-driving cars). It is in one's interest to collect as many labeled examples as possible in such a setting. How useful is self-supervision in these cases? How do we measure this utility?

Given a downstream task, several outcomes are possible when comparing a finetuned self-supervised model against a baseline trained from scratch (illustrated in Fig.~\ref{fig:intro_fig}). With more labeled data, the performance of a model will improve and may eventually plateau. But in practice, one has a finite labeling budget, and that budget will determine the accuracy reached when training from scratch. There are three subsequent outcomes for the finetuned model: (a) self-supervision achieves a better accuracy than the baseline; (b) self-supervision achieves the same accuracy but with fewer labeled examples; (c) self-supervision achieves the same accuracy with the same number of labeled examples.

For each of these outcomes, we can quantify the utility of self-supervision as the saving in labels. That is, to achieve the same accuracy without self-supervision, how many more labels would be needed. Specifically, if we define $a(n)$ as the accuracy of a model trained from scratch given $n$ labeled examples and $a_{ft}(n)$ as the accuracy of the finetuned model, the utility at $n$ is defined as $U(n) = \hat{n} / n - 1$ where $\hat{n}$ is the number of labels required such that $a(\hat{n})=a_{ft}(n)$. This is the ratio of additional labels needed to match the accuracy of the finetuned model. The utility is zero when self-supervision achieves the same maximum accuracy without lowering labeling cost ($\hat{n}=n$). The utility is infinite when there does not exist any number of labeled samples such that the model trained from scratch matches the finetuned model.

Note that one might not expect self-supervision to help in the presence of many labeled examples, because in the limit the entire input space would be densely covered, and a deep network just needs to fit the labeled  data well. However, this is based on the false assumption that we can fit large labeled data arbitrarily well. SGD training is not guaranteed to reach a global optimum, and self-supervised pretraining may produce better representations that help optimization, just as residual links improve fitting to the data.

Given the above definition of utility, we systematically evaluate a number of recent self-supervised algorithms. To do this, we construct a benchmark of synthetic images. A synthetic benchmark offers unique advantages. It allows easy generation of a large number of labeled examples. It also allows easy exploration of a variety of downstream tasks from classification to dense prediction and from semantics to geometry. Finally, it allows precise control of the complexity of the data and the difficulty of the downstream task through factors such as color, texture, and viewpoint. Our main contribution is a thorough exploration along all of these dimensions to provide insights into where and when one can expect self-supervision to be useful in practice.

We find that leading self-supervised pretraining methods are useful with a small labeling budget, but utility tends to decrease with ample labels. In particular, as the number of labels increases, the most common outcome is Fig. \ref{fig:intro_fig} (c), where gains from self-supervised pretraining tend to diminish before performance plateaus for the training-from-scratch baseline. We also find that self-supervision is more helpful when applied to larger models and to more difficult versions of the data. Moreover, we find that relative performance of methods is not consistent across downstream settings, and that the commonly used linear evaluation does not correlate with utility.


\section{Related Work}

Pretraining has long been used to improve performance on visual tasks \cite{donahue2013decaf, girshick2014rich}. The features learned by a convolutional network trained on a large dataset like ImageNet \cite{deng2009imagenet} transfer well to many settings. It has further been shown that pretraining is effective given exceedingly large, noisy or weakly-labeled datasets \cite{sun2017revisiting, mahajan2018exploring, yalniz2019billion, kolesnikov2019large}. In this work we focus specifically on the results that can be achieved with self-supervised pretraining, so do not consider the use of weak labels nor do we touch on semi-supervised methods that pseudolabel data \cite{berthelot2019mixmatch, xie2019self, yalniz2019billion}.

Recent work has shown that with sufficient training time, a model trained from scratch can match ImageNet-pretrained performance on COCO \cite{he2019rethinking}. We do not investigate how to improve training from scratch to match pretraining, nor do we benchmark ImageNet pretraining. ImageNet pretraining is subject to concerns around domain shifts, so self-supervised methods have an advantage in our benchmark since pretraining can be performed on the exact image distribution used for the downstream task.

\begin{figure*}[t]
\begin{center}
   \includegraphics[width=\linewidth,trim=0 0 0 130,clip]{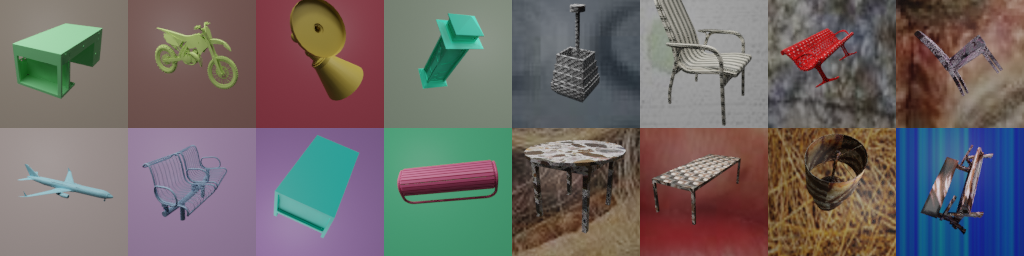}
\end{center}
\vspace{-.1cm}
\caption{Example images from four datasets of increasing complexity (from left to right) controlling for viewpoint and texture.}
\label{fig:ex_syn_imgs}
\vspace{-.2cm}
\end{figure*}

The past several years have seen a wide variety of methods proposed for visual self-supervision \cite{jing2019self, doersch2017multi, gidaris2018unsupervised, caron2018deep, bojanowski2017unsupervised, zhuang2019local}. These methods can take many forms and also rely on a variety of cues such as information across frames in video \cite{zhou2017unsupervised, wang2015unsupervised} or across different image modalities \cite{tian2019contrastive, sayed2018cross, ren2018cross}. One family of methods are those that center around reconstruction. This often takes the form of an autoencoder \cite{hinton2006reducing, kingma2013auto, zhang2017split} or alternatively requires inferring some missing part of the data by performing tasks like inpainting \cite{pathak2016context} or colorization \cite{zhang2016colorful, vondrick2018tracking}. Other methods take advantage of spatial properties of images. This can include judging the relative spatial position of image patches \cite{doersch2015unsupervised, noroozi2016unsupervised}, or predicting what sorts of transformations have been applied to an image \cite{gidaris2018unsupervised, dosovitskiy2014discriminative}.

Recently, a wave of methods based on contrastive embeddings have proven effective for pretraining \cite{oord2018representation, henaff2019data, tian2019contrastive, wu2018unsupervised, bachman2019learning, ye2019unsupervised}.  These methods produce features that maximize mutual information either across representations at different spatial locations of the image \cite{henaff2019data,bachman2019learning} or across different views \cite{tian2019contrastive} of the image. The key idea being that different patches of an image or different versions of the image (\eg different image channels or versions augmented with data transformations) should map to a similar embedding that is unique across image samples.

There have also been a number of recent undertakings to provide comprehensive evaluation of today's self-supervised methods \cite{oliver2018realistic, goyal2019scaling, locatello2018challenging,zhai2019visual,kolesnikov2019revisiting,doersch2017multi,resnick2019probing}. Each work focuses on distinct facets such as how the choice of architecture design affects performance \cite{kolesnikov2019revisiting} or to what degree methods produce disentangled representations and how this affects downstream task performance \cite{locatello2018challenging}. Other work has measured performance by scaling up the number of unlabeled images and increasing the difficulty of the self-supervised tasks \cite{goyal2019scaling}. A distinguishing feature of our work is an emphasis on measuring utility both with large numbers of labels and when finetuning the full model. Moreover, our synthetic setting provides the opportunity to gain insights into how image complexity affects self-supervision performance.


\section{Self-supervised Pretraining}

We follow a fixed strategy for pretraining and finetuning. During pretraining, a self-supervised algorithm is chosen, and the model is presented with unlabeled images to fit the specified loss. During finetuning, a new output layer is added to the network for a target downstream task and the model is trained on labeled images to fit the task as well as possible. At no point is the network jointly trained on both the self-supervised task and downstream task, both for simplicity and to reflect typical use of pretrained models.

The usefulness of self-supervised pretraining depends on a number of factors. We break down these factors across the following four categories:

\begin{itemize}
    \item \textbf{Data:} There is a close connection between the difficulty of a dataset and the number of labeled examples required to saturate performance on a task. Thus it is important to run experiments that control for not just the number of images but their complexity. Here, complexity refers to factors of variation such as those that arise due to changes in lighting, texture, and viewpoint.

    \item \textbf{Model:} The highest level of performance possible on a task depends on the model used to learn that task. We do our best to control for the backbone model such that a fair comparison is made between methods.

    \item \textbf{Self-supervision algorithm:} Self-supervised algorithms rely on different cues for learning, and this choice may affect downstream performance. For example, a particular method may explicitly train a model to be invariant to features required by a particular downstream task.

    \item \textbf{Downstream task:} Similarly, different tasks may lend themselves better or worse to different types of pretraining. Moreover, the difficulty of the downstream task will affect how much data is needed to do well and the degree to which performance plateaus as the number of labels increases.
\end{itemize}

\begin{figure*}[t]
\begin{center}
   \includegraphics[width=\linewidth]{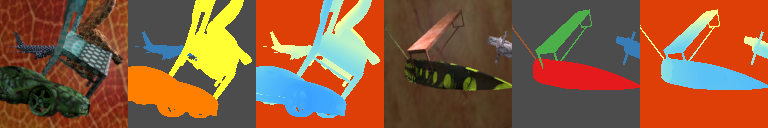}
\end{center}
\vspace{-.1cm}
   \caption{Example images in the multi-object setting as well as the ground truth semantic segmentation and depth.}
\label{fig:ex_multiobj_imgs}
\vspace{-.2cm}
\end{figure*}

To summarize, in this work we will be assessing the effectiveness of pretraining with different data, different backbones, different pretraining, and different downstream tasks. The interplay of all of these factors is complicated, so it is important to see how they affect each other under as many settings as possible. For a particular setting, we will pretrain a model and then finetune it while varying the amount of available labels. We compare performance to a baseline model trained from scratch.


\section{Synthetic Benchmark}

The need to control for dataset difficulty as well as the need for many labeled images motivates our use of synthetic images. We can control all factors of the generation process and produce an endless supply of images. Moreover, it is trivial to acquire annotations for a wide variety of tasks - many of which would be difficult or impossible to collect for real world images such as ground truth depth.

Our synthetic images consist of objects floating in empty space. For a given image, we can change the number of objects, their orientation, their texture, as well as the lighting conditions of the scene. If the scene only consists of a single object, we keep the position of the object fixed in the center of the image at a fixed distance. We also normalize the scale so that the size of objects is consistent across classes and models. If there are multiple objects, their positions are chosen randomly such that they are evenly dispersed and remain mostly in the camera frame.

We render images with Blender \cite{blender} using object models from ShapeNet \cite{shapenet2015}. We choose 10 object classes to use in all versions of the synthetic data (\emph{airplane, bench, cabinet, car, chair, lamp, sofa, table, watercraft, motorcycle}). To increase image diversity and account for classes with a limited number of models, we augment the models with occasional random stretching along spatial axes. For all datasets, a strict split is enforced across our training and evaluation settings. A random subset of 80\% of the models are used for training, the remaining 20\% are used during validation and testing. For consistency, this same split is enforced across all versions of our synthetic data.

\subsection{Factors of variation}

When rendering, we control for four different sources of image variation. The complexity of the dataset increases as more sources of variation are used.

\textbf{Texture:} We either apply a flat color material or a texture to the ShapeNet objects and background (as seen in Figure \ref{fig:ex_syn_imgs}). Textures are taken from DTD \cite{cimpoi14describing} which provides a wide variety of images to source from. These can vary from basic patterns such as colored stripes to photos of intricate textures found in the world.

\textbf{Color:} In the real world, object classes are often associated with colors, and many self-supervised techniques rely on this fact to train models \cite{zhang2016colorful,tian2019contrastive}. With this in mind, we define two options for using color in the synthetic data. In the easier setting, each class is associated with a fixed color distribution. Concretely, a random hue is assigned to each class. When rendering an object, a new color is chosen by sampling from a normal distribution around the corresponding hue for that class. We follow this strategy even when a texture is applied to the object by mixing the texture image with the target color.

It is not trivial to predict the object class strictly from the color of pixels in the image as the distribution of colors will overlap across classes. But still, the correlation between color and object class does make classification easier. In the harder setting, each object is rendered with a random color so that there is no correlation whatsoever between the color and object class.

\textbf{Viewpoint:} We render objects at either a fixed or random orientation. At the fixed orientation, all objects are viewed such that there are clearly visible features to distinguish each class as opposed to an ambiguous overhead or head-on view. When randomly sampling viewpoints, a rotation is chosen from a normal distribution with a mean at the previously described fixed orientation. The deviation is large enough to include many extreme viewpoints, but views are more likely to be sampled close to the original fixed viewpoint.

\textbf{Lighting:} There is a single light source in the scene. We either render data with the the light in a fixed position or randomly placed for each sample.


\section{Downstream Tasks}

To ensure we evaluate on a variety of downstream tasks, we consider common distinguishing features of computer vision problems. For example, tasks are often differentiated by whether they pertain more to semantic or geometric information. An example of the former would be recognizing object categories, while an example of the latter would be surface normal estimation. Another major distinction amongst vision tasks is whether or not a task requires dense predictions across the space of the image. This is the difference between object classification and semantic segmentation where classification is done per pixel.

It is important to consider these distinctions as there could be differences in whether pretraining is relevant for geometric as opposed to semantic features. Furthermore, many pretraining methods are designed around producing a single feature vector, so it is unclear how well these methods serve for downstream tasks that require dense prediction.

The tasks that we benchmark on are object classification, object pose estimation, semantic segmentation, and depth estimation. These tasks provide a contrast between extracting semantic and geometric information (classification/segmentation vs. pose/depth), as well as a contrast between predicting global or dense features (classification/pose vs. segmentation/depth).

Another factor guiding our choice of tasks is simplicity. We benchmark on tasks that do not require sophisticated losses, two-stage pipelines, or complicated post-processing. This is a practical decision to limit the hyperparameters that would govern training behavior and allow us to focus more directly on the impact of pretraining.

\subsection{Task Details}
\label{task_details}

\noindent\textbf{Object classification:} Object classification is one of the most standard benchmarks for evaluating self-supervised pretraining. For this task, we train the model to distinguish between the ten ShapeNet classes used to render our synthetic data. Images are generated such that they only contain a single object, and a uniform distribution exists across all ten classes. We measure performance by standard classification accuracy.

\smallskip\noindent\textbf{Object pose estimation:} To evaluate object pose estimation, again we use images that only contain a single, centered object. Rather than predicting a full rotation matrix (or some alternative representation) we discretize pose into five bins and train a classifier.

One reason for framing the problem this way is to account for the rotational symmetry present in some of the ShapeNet categories (namely lamps and tables). The five bins are chosen such that orientation along that rotational axis is ignored. The model must predict whether the top face of the object is either oriented upwards, forwards, backward, to the left, or to the right.

This formulation for pose estimation still requires that the model extract features relevant for 3D understanding, but alleviates some of the complications that arise when supervising and evaluating pose. The bins are chosen such that samples are evenly distributed across all five categories. We train with a cross entropy loss and report classification accuracy.

\smallskip\noindent\textbf{Semantic segmentation:} For semantic segmentation, images are rendered with multiple objects (as seen in Figure \ref{fig:ex_multiobj_imgs}). We are not concerned with designing a model to output high resolution or precise segmentation masks, so instead we supervise at a much coarser resolution relative to the input image. We apply a per-pixel cross entropy loss and report average classification accuracy.

\smallskip\noindent\textbf{Depth estimation:} Just as with semantic segmentation, depth estimation is tested on images with multiple objects and at a much coarser resolution than the original input image. We supervise depth with an L1 loss, and measure accuracy with a standard metric in the literature ($\delta < 1.25$) that measures the percentage of predictions that fall within a given ratio of the ground truth depth.

\begin{figure*}[t]
\begin{center}
   \includegraphics[width=\linewidth]{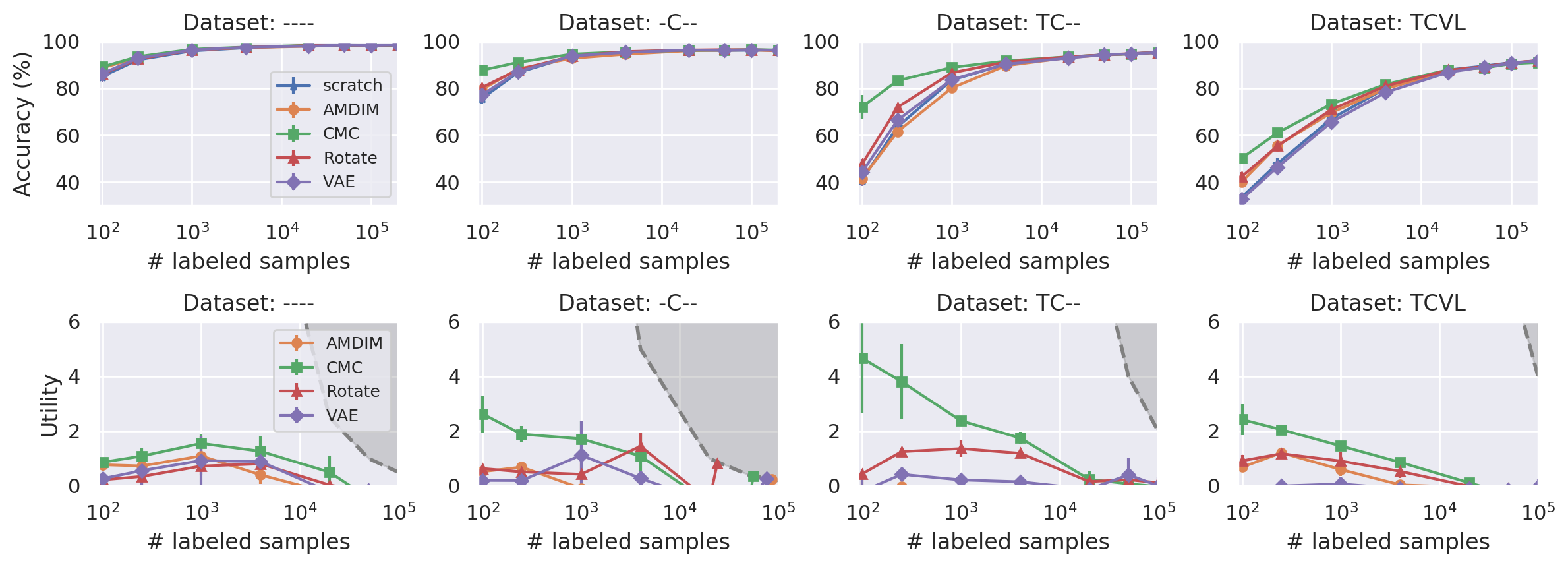}
   \caption{Object classification accuracy and utility of pretrained ResNet9 models when finetuning on increasing numbers of labeled samples. As more labeled data is included, the utility (ratio of labels saved) tends toward zero eventually converging with performance when trained from scratch. This occurs before model performance has saturated.}
    \label{fig:cls_acc}
\end{center}
\vspace{-.2cm}
\end{figure*}


\section{Pretraining Methods}

We choose four different self-supervised algorithms for pretraining:
\begin{itemize}
    \item \emph{Variational autoencoder (VAE)} \cite{kingma2013auto}: A standard, established baseline for mapping images to a low-dimensional latent space.
    \item \emph{Rotation} \cite{gidaris2018unsupervised}: A simple yet effective method for pretraining. The network is tasked with predicting whether an image has been rotated either 0, 90, 180, or 270 degrees.
    \item \emph{Contrastive Multiview Coding (CMC)} \cite{tian2019contrastive}: A recent method for self-supervision that works by splitting an image into multiple channels such as the \emph{L} and \emph{ab} channels of an image in Lab color space. The separated channels are passed through two halved networks and the output embeddings are compared and contrasted to embeddings from other images.
    \item \emph{Augmented Multiscale Deep InfoMax (AMDIM)} \cite{bachman2019learning}: Similar to CMC, this method also trains a model through contrastive coding. Instead of comparing across image channels, AMDIM compares representations from two augmented versions of the same image as well as representations produced at intermediate layers of the network.
\end{itemize}

These methods were chosen to strike a balance between styles of self-supervision as well as complexity of the self-supervised task. Rotation, CMC, and AMDIM are three high performing self-supervision methods as measured by pretraining and evaluation on ImageNet.


\section{Experiment Details}

\noindent\textbf{Datasets:} We render 15 dataset variations for our experiments. The key distinguishing feature of the datasets is whether they belong to the lower-resolution single-object setting or higher-resolution multi-object setting. We use single-object images to evaluate object classification and pose estimation, and multi-object images for semantic segmentation and depth. The dataset resolutions are 64x64 and 128x128 respectively.

For most datasets we render 240,000 images. The only exception are the single-object datasets with viewpoint changes where we render 480,000 images total. A subset of 15\% of the images is held out for validation and testing.

In the figures presented in this paper we use a shorthand to summarize the factors of variation of a particular dataset. Each letter corresponds to a particular factor, and a dash (-) signals that the easier version of that factor is used. To summarize, T: \emph{Texture} (flat colors vs DTD textures); C: \emph{Color} (fixed distribution vs random colors); V: \emph{Viewpoint} (fixed viewpoint vs random viewpoint); L: \emph{Lighting} (fixed lighting vs random lighting). An example dataset would be `TC-{}-' where DTD textures are applied along with random colors but viewpoint and lighting are fixed.

\noindent\textbf{Models:} We use ResNet9 \cite{page2019resnet} and ResNet50 \cite{he2015deep} for all experiments. The ResNet9 model suits our experiments well since it is much faster to train and converge, while ResNet50 is more commonly used and can illustrate how results change with additional network capacity.

We reduce the amount of pooling performed by the models to account for the fact that we are benchmarking on lower resolutions than typical for ResNet models. For the dense prediction tasks, we predict outputs from the features just before global pooling. Thus we supervise at a resolution of 16x16. While this is low, it allows us to assess the model's ability to produce features for dense tasks with zero modifications to the network backbone.

\noindent\textbf{Training:} For a given dataset, we pretrain all self-supervised algorithms on all available training images. The self-supervised algorithms are trained for between 100-200 epochs depending on the algorithm. For finetuning, we load a pretrained model and train for between 75 to 200 additional epochs. This is determined by the amount of images used, and whether we are performing dense prediction.

\noindent\textbf{Evaluation:} On all tasks we report standard performance metrics as described in Section \ref{task_details}. We measure how performance changes as more labeled examples are provided during finetuning. Furthermore, we report the utility measure as described in the introduction, which measures the proportion of additional labels that would be necessary to reach the same accuracy without pretraining. For example, if a self-supervised algorithm reaches a certain accuracy at 100 samples, and it requires 500 samples to reach that performance when training from scratch, then $U(100) = (500/100) - 1 = 4$. We report the change in utility as a function of labeled samples for each self-supervised algorithm.

\begin{figure*}[t]
\begin{center}
   \includegraphics[width=\linewidth]{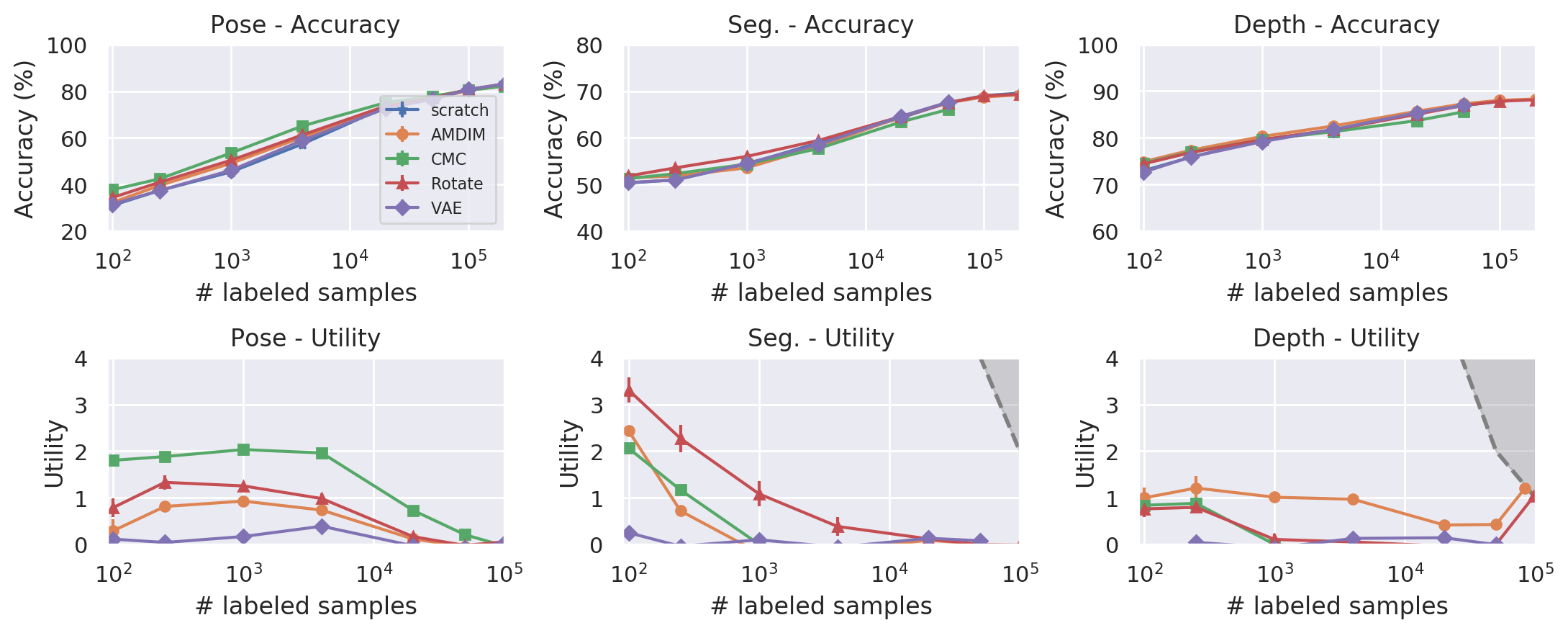}
\end{center}
\vspace{-.1cm}
   \caption{Performance on additional downstream tasks with ResNet9 on the hardest dataset setting (TCVL). The best performing method differs depending on the downstream task suggesting that diverse settings should be considered when comparing self-supervised models.}
\label{fig:pose_acc}
\vspace{-.2cm}
\end{figure*}

\begin{figure}[t]
\begin{center}
   \includegraphics[width=\linewidth]{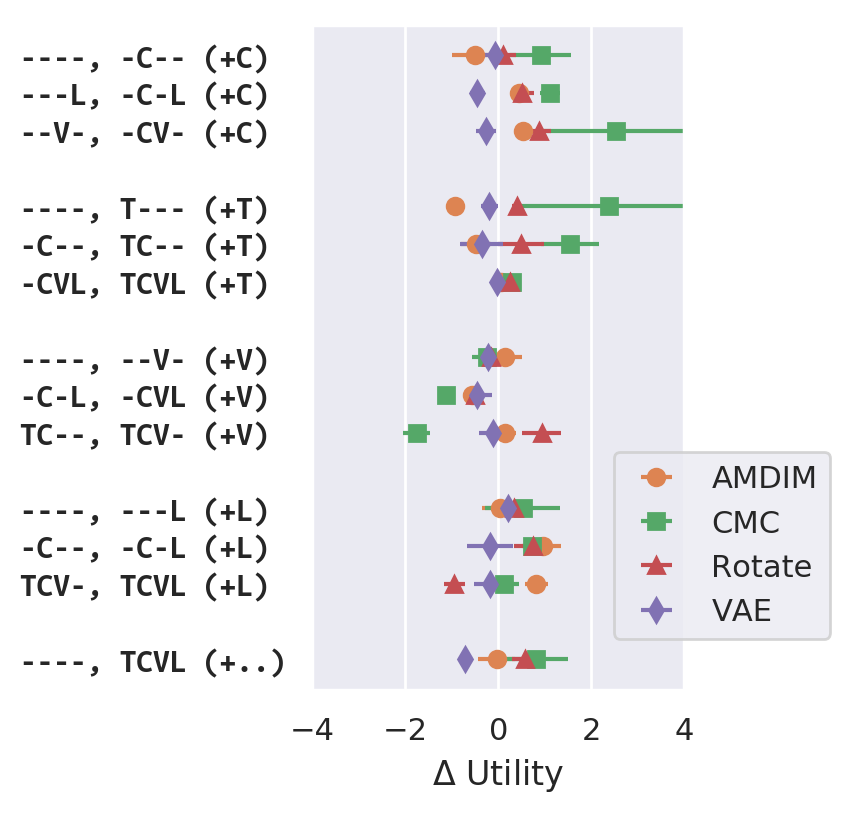}
\end{center}
\vspace{-.1cm}
   \caption{Change in utility across datasets when controlling for factors of image variation (all models trained with ResNet9). The factors are: color (C), texture (T), viewpoint (V), and lighting (L).}
\label{fig:utility_compare}
\vspace{-.2cm}
\end{figure}

\begin{figure*}[t]
\begin{center}
   \includegraphics[width=\linewidth]{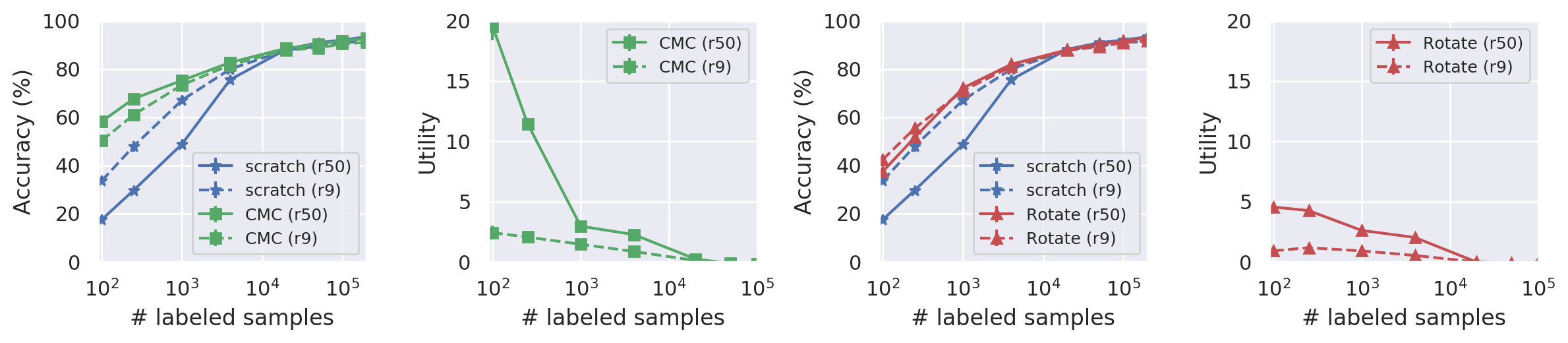}
   \caption{Comparison between ResNet9 and ResNet50 backbones for object classification on TCVL. With few labeled samples the performance of the ResNet50 model is worse when trained from scratch, but when pretrained is better than the pretrained ResNet9 suggesting the importance of pretraining large models when working with less data.}
    \label{fig:r9_v_r50}
\vspace{-.3cm}
\end{center}
\end{figure*}

Note that utility cannot be calculated if a model reaches accuracies higher than the maximum reached by the baseline trained from scratch. This does not mean that there does not exist a corresponding value since it is possible the baseline may reach higher accuracy given a larger budget, but does mean that we cannot compute a utility value appropriately. To visualize this we gray out areas in which utility cannot be calculated. Concretely, if the maximum accuracy of training from scratch is reached at 100k samples, then when calculating $U(50\text{k})$ no value above 1 could be computed and thus this area would be grayed out.


\section{Results}

\noindent\textbf{Utility vs Number of Labeled Samples:} 
We first study how utility changes as the number of downstream training examples increases. For simplicity, we start with results on object classification shown in Figure \ref{fig:cls_acc}.

We see that self-supervision has significant utility when the number of labeled samples is small, but utility approaches zero as labeled data grows. This observation holds for all pretraining methods across downstream settings. Performance of the pretrained models converges with performance of the baseline before accuracy of the model plateaus on the task. This suggests that the utility of self-supervised pretraining comes mainly from better regularization that reduces overfitting, not better optimization that reduces underfitting---otherwise we should expect self-supervision to have non-negligible utility even with large numbers of labeled samples.

\noindent\textbf{Utility vs Downstream Task:}
Next, we investigate whether pretraining algorithms are more or less useful across different downstream tasks. In Figure \ref{fig:pose_acc} we report performance on pose estimation, depth, and semantic segmentation. For the sake of space we only report performance on one dataset variation (TCVL), but more results can be found in the Appendix.

The relative rank of different pretraining methods changes with the choice of downstream task. Where CMC performs best in object classification and object pose estimation, we observe that rotation and AMDIM perform better on segmentation and depth estimation respectively.

Because utility depends on the downstream setting, an important implication is that object classification performance may not be predictive of performance on other tasks. It is thus important to consider diverse downstream settings when evaluating self-supervised methods, and the best pretraining method for a practitioner will depend on the specific context in which they wish to use their model.

\begin{figure*}[t]
\begin{center}
   \includegraphics[width=.28\linewidth]{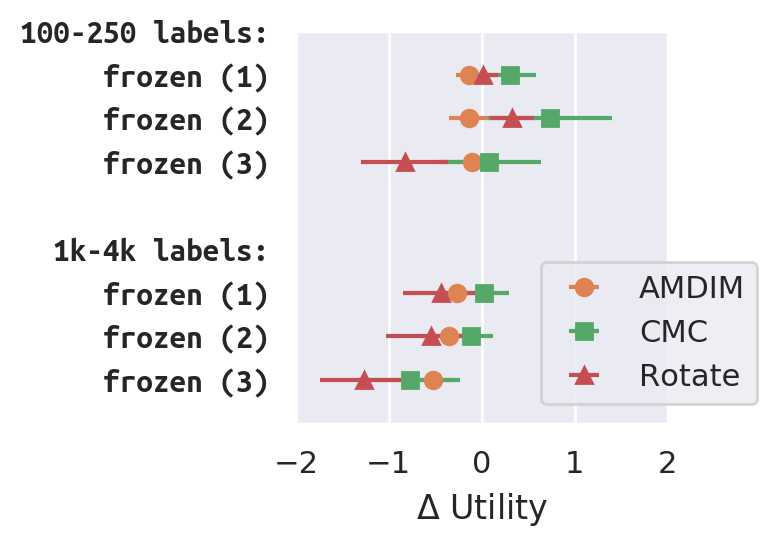}
   \includegraphics[width=.71\linewidth]{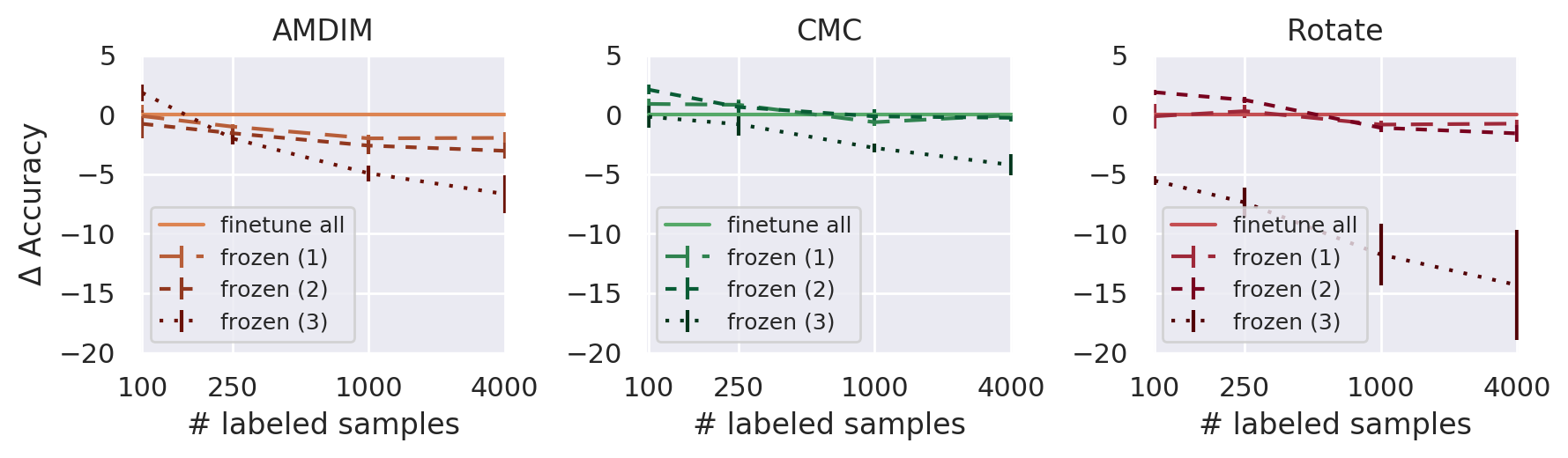}
\end{center}
   \caption{Finetuning performance when freezing different amounts of the network. Change in utility and accuracy is reported relative to a baseline where all weights are finetuned. The numbers \emph{(1-3)} indicate the number of blocks of the ResNet model that have been frozen.}
\label{fig:frozen}
\vspace{-.4cm}
\end{figure*}

\noindent\textbf{Utility vs Data Complexity:}
Our experiments also allow us to measure how factors of image variation impact the utility of different pretraining methods. In Figure \ref{fig:utility_compare}, we report the change in utility for each method across dataset pairs controlling for individual factors.

We observe relatively consistent changes to the utility of a particular algorithm when adjusting a given factor of image variation. For example, upon introducing random colors or textures, the utility of CMC consistently goes up, but the addition of viewpoint changes results in a drop in utility. These effects occur across multiple dataset pairs.

Changes to utility for each factor differ across pretraining algorithms. That is, where utility goes up with texture and down with viewpoint changes for CMC, the opposite is true for AMDIM. One possible source of this discrepancy is that as opposed to CMC that applies a loss to global features, AMDIM encourages intermediate local features across the image to map to similar embeddings. This may increase sensitivity to evidence in local windows, which would explain the adverse effect of random textures and robustness to viewpoint changes where local object evidence changes less than the global arrangement of object parts.

For VAE, additional factors of variation lower utility. A possible explanation is that as a reconstruction-based approach, the latent space must encode all information necessary to reproduce the image. As data complexity increases, more spurious details must be captured that do not pertain to the downstream task. Contrastive approaches on the other hand, teach a network to map to the same embedding after applying different image transformations. Thus the network learns to ignore changes in pixel space that do not correspond to changes in semantic object class that a VAE would otherwise encode.

\begin{table}
\centering
\small
\resizebox{\linewidth}{!}{
\begin{tabular}{lcccccccc}
  & ---- & -C-- & -C-L & -CVL & T--- & T-VL & TC-- & TCVL \\ \hline
normal training & 98.5 & 96.2 & 93.8 & 92.1 & 97.4 & 91.8 & 95.3 & 91.5 \\ \hline
AMDIM & 96.9 & 94.9 & 89.8 & 77.4 & 92.5 & 73.0 & 80.4 & 74.5 \\
CMC & 88.7 & 88.9 & 87.5 & 72.5 & 91.3 & 69.9 & 90.8 & 69.0 \\
Rotate & 31.4 & 30.0 & 39.7 & 30.6 & 32.2 & 37.6 & 25.4 & 32.9 \\
VAE & 71.7 & 63.8 & 51.0 & 35.7 & 72.8 & 26.9 & 36.8 & 31.5 \\ \end{tabular}
}
\vspace{.2cm}
\caption{Results of linear evaluation for object classification.}
\label{tab:linear}
\vspace{-.4cm}
\end{table}

\noindent\textbf{Linear evaluation:} 
In Table \ref{tab:linear} we report performance of each self-supervised method using the linear evaluation common in other work. We freeze each pretrained model and train a single linear layer for object classification. Note, the linear layer is trained with all available samples for that dataset. We report performance across eight datasets for object classification with a ResNet9 model.

The linear evaluation results do not reflect finetuned performance shown in Figures \ref{fig:cls_acc}. Despite AMDIM having the best linear performance, its utility is consistently lower than both CMC and Rotation. While linear evaluation is informative for leveraging a frozen model, it may not correspond to which models perform best when finetuned.

\noindent\textbf{Utility vs Model Size:} Next, to measure the effect of model capacity, we provide a comparison between ResNet9 and ResNet50 performance on object classification in Figure \ref{fig:r9_v_r50}.

Utility is always higher with a ResNet50 backbone. This is due to both a performance drop of the baseline and performance improvement of the finetuned model when transitioning to the larger model. The baseline drop occurs at small dataset sizes and ResNet50 does outperform the ResNet9 given sufficient labels.

CMC achieves higher performance with the larger backbone even when finetuning with less labels. This suggests that for best downstream performance, it is helpful to pretrain on as large a backbone as possible. Similar findings are shown when pretraining with noisy labels \cite{kolesnikov2019large}.

\noindent\textbf{Utility vs Amount of Finetuning:}
To further expand on the comparison between finetuning and the linear evaluation, we measure performance after finetuning a network frozen up to different intermediate layers. The ResNet9 model is made up of three main blocks of layers, so we test how finetuning performance changes if we freeze the model up to each block. Note that after the third block, there are two fully-connected layers so this is slightly more expressive than the linear evaluation baseline. In Figure \ref{fig:frozen} we show both the change in utility and change in accuracy after freezing increasing amounts of the network. Results are averaged across multiple dataset variations.

Performance suffers as more of the model is frozen. As expected, when most of the network is frozen (`frozen (3)') the pretraining techniques that have the highest linear performance see less of a drop in performance. And though some freezing is helpful when the number of labels is low (100-250 samples), the best accuracy is consistently reached by models that are fully finetuned.


\section{Conclusion}

In this work we investigate a number of factors that affect the utility of self-supervised pretraining. We provide a thorough set of experiments across different downstream tasks and synthetic datasets to measure the utility of pretraining with state-of-the-art self-supervised algorithms. Our study shows that the greatest benefits of pretraining are currently in low data regimes, and utility approaches zero before performance plateaus on the task from additional labels. Further, performance of a self-supervised algorithm in one setting may not necessarily reflect its performance in others, underscoring the importance of studying and evaluating pretraining methods across diverse scenarios.

\paragraph{Acknowledgements} The authors would like to thank Jonathan Stroud for his feedback on paper drafts and for many helpful discussions. This work was partially supported by the National Science Foundation under Grants No. 1734266 and No. 1617767.

\clearpage

{\small
\bibliographystyle{ieee_fullname}
\bibliography{egbib}
}

\clearpage

\section*{Appendix}

\appendix

\section{Additional training details}

We use the Ray Tune library \cite{moritz2018ray,liaw2018tune} for selecting hyperparameters. We tune the learning rate schedule and weight decay terms as well as task-specific parameters in the self-supervised settings. The pretrained models are given as much training time during finetuning as the model trained from scratch. Precise details and experiment configuration settings can all be found in the released code.

We apply standard data augmentation (cropping, flipping, brightness/color perturbations, Cutout \cite{devries2017improved}) in all settings with a few exceptions for specific tasks. For example, to avoid issues that would arise with padding and resizing for depth estimation we do not do random cropping.

One additional note, since Contrastive Multiview Coding (CMC) \cite{tian2019contrastive} uses Lab color space as input, we present all images across all methods and datasets in Lab color space for consistency. In ablations, we do not find this affects performance when training a supervised model from scratch with a large amount of labeled data, but want to control for as many differences as possible between methods.

\section{Additional experiments}

\subsection{Cross-dataset transfer}
\begin{figure}[t]
\begin{center}
   \includegraphics[width=.9\linewidth]{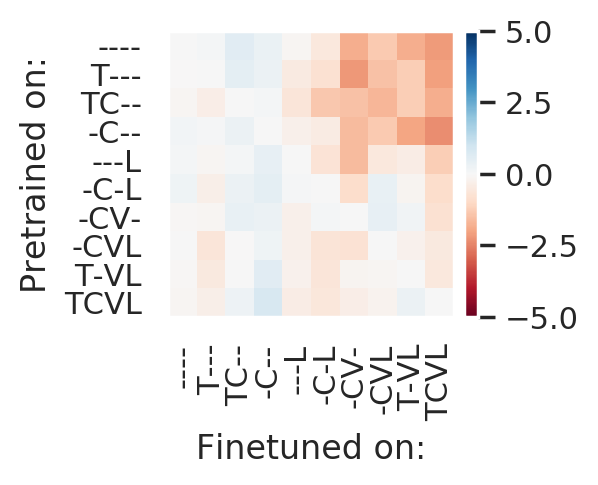}
\end{center}
\vspace{-.2cm}
   \caption{Change in performance when pretraining and finetuning on different dataset variations. Report average change in accuracy when finetuning on 250-4000 samples for models pretrained with AMDIM.}
\label{fig:transfer}
\vspace{-.3cm}
\end{figure}

A common concern with self-supervised pretraining is robustness to domain shifts. To get a sense of how shifts in dataset properties affect performance, we compare models pretrained and finetuned on different datasets. We evaluate downstream object classification performance in the low data regime. Results are shown in Figure \ref{fig:transfer}.

Exposure to viewpoint changes appears to affect finetuning performance most. Accounting for all other image factors, models that are pretrained on datasets without viewpoint changes and finetuned on datasets with viewpoint changes suffer the most consistent drop in performance (upper right of Figure \ref{fig:transfer}).

\begin{figure*}[t]
\begin{center}
   \includegraphics[width=\linewidth]{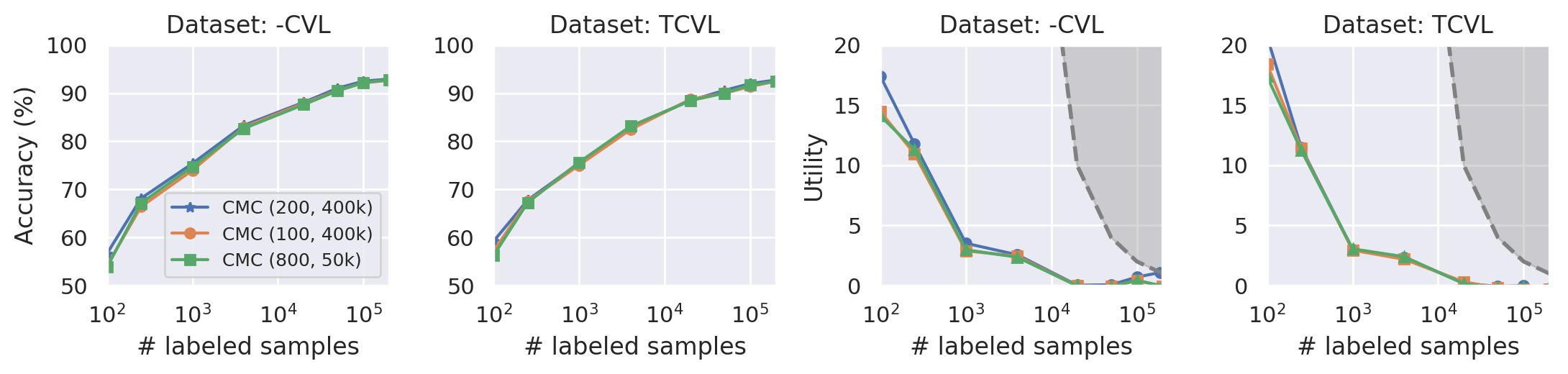}
\end{center}
\vspace{-.2cm}
   \caption{Change in utility with different pretraining settings for CMC. We label each pretrained model with (number of pretraining epochs, number of unlabeled images) and report downstream object classification performance. Because 50k unlabeled images is 1/8th the base 400k, the model is trained for more epochs to match total number of training iterations.}
\label{fig:ablate_cmc}
\vspace{-.1cm}
\end{figure*}

\begin{figure*}
\begin{center}
   \includegraphics[width=\linewidth]{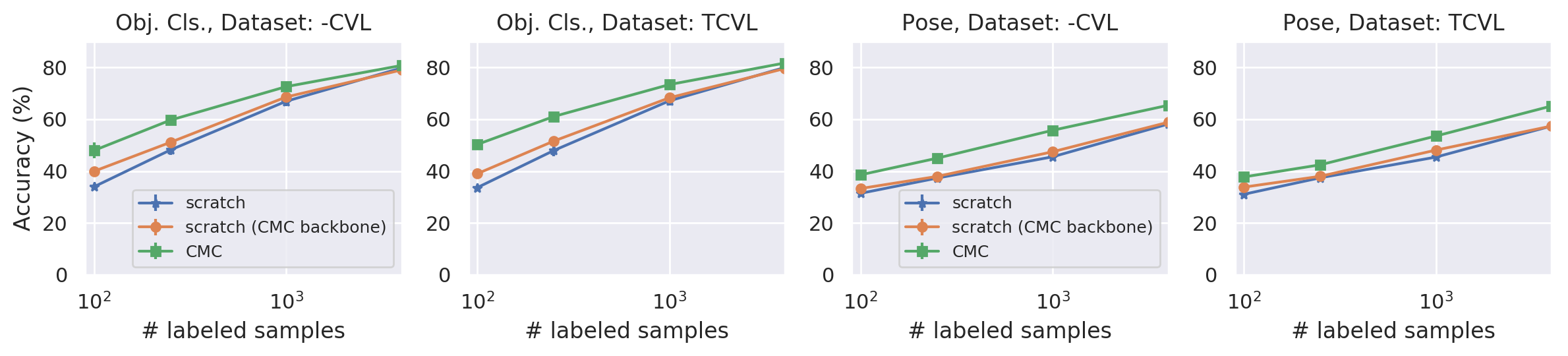}
\end{center}
\vspace{-.3cm}
   \caption{Training from scratch with CMC backbone (ResNet9 base) on object classification and object pose estimation.}
\label{fig:cmc_half_cls}
\begin{center}
   \includegraphics[width=\linewidth,trim=0 0 0 0,clip]{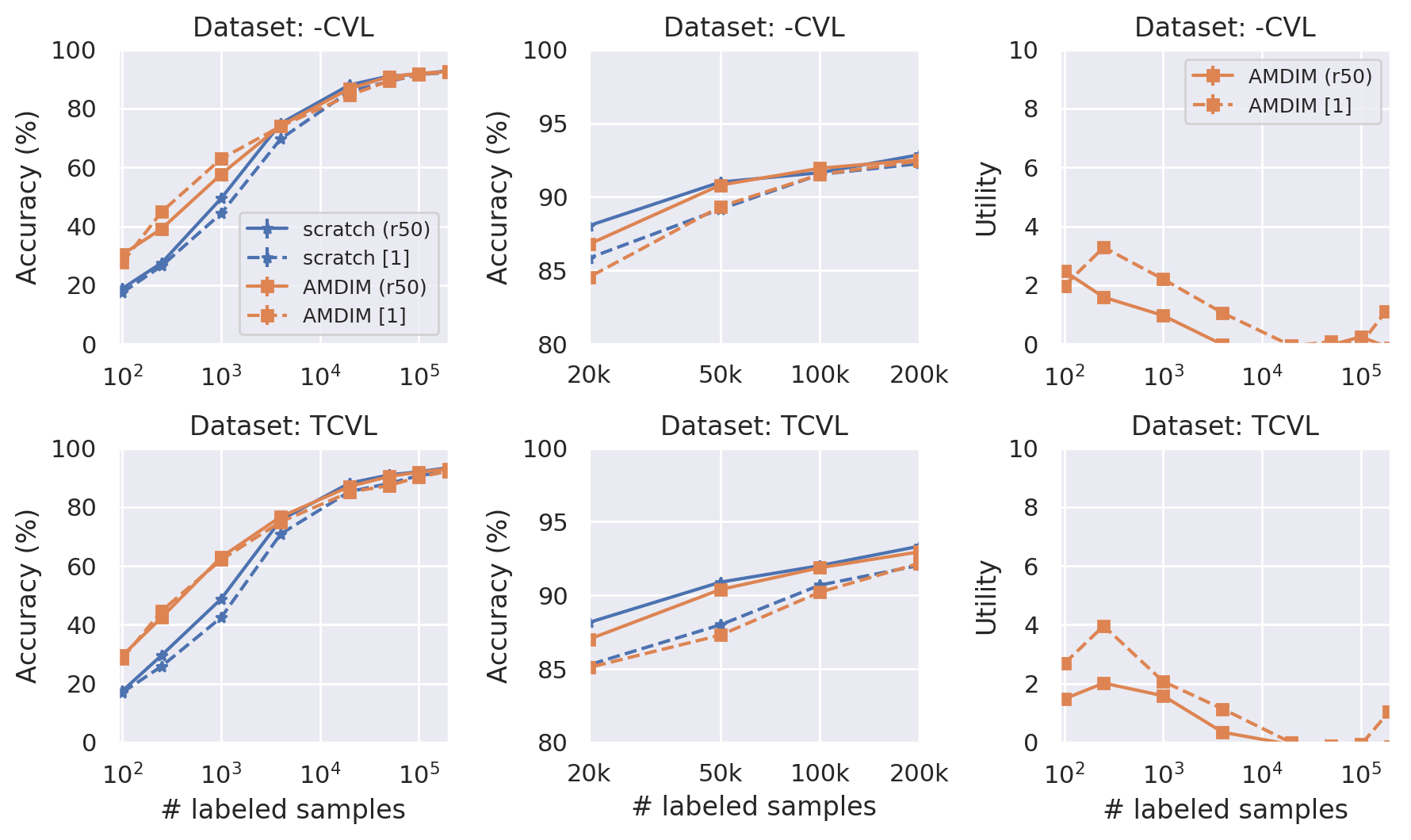}
\end{center}
\vspace{-.3cm}
   \caption{Comparison between ResNet50 backbone and backbone proposed in AMDIM paper \cite{bachman2019learning}. Number of layers and number of channels are chosen to control for total parameters so that the two models are matched in size.}
\label{fig:depth_r9_r50}
\end{figure*}

\subsection{Tuning the pretraining process} Without a quick proxy for downstream performance, it is difficult to tune and improve the pretraining process. As seen in Figure \ref{fig:ablate_cmc}, the downstream effect of more unlabeled images and more training time may be limited. We observe either no effect or a modest adverse effect when training for half the time or with an eighth of the unlabeled images. In practice, hyperparameter tuning is challenging as it is expensive to evaluate the impact, if any, of further adjustments to the pretraining process, and exponential increases in pretraining time and data might result in little to no change on the final task.

\subsection{Network backbones}

\noindent\textbf{CMC:} As noted in the paper, the backbone is different when pretraining with CMC. This is because the authors split the network in half, where each half is responsible for either processing the L channel or the ab channels of the image. This results in a restricted network with approximately half the original number of parameters. In all other ways, the network is identical.

This leads to a modest change in performance in single object settings (Figures \ref{fig:cmc_half_cls}). When evaluating utility in the paper we measure relative to the original baseline of the full model trained from scratch. This does not affect the main conclusions of the paper that utility approaches zero as the number of labeled samples increases and that relative performance depends on the downstream setting.

\noindent\textbf{AMDIM:} In the paper proposing AMDIM \cite{bachman2019learning}, a different backbone is used and recommended. We compare performance between this backbone and a ResNet50 model while trying to control for overall model size as much as possible (similar feature activation sizes and total parameter count). The linear performance of the proposed model is much better than that of the ResNet50, but when finetuned, the ResNet50 achieves higher accuracy. This is especially true at larger dataset sizes. For the purposes of the comparisons made in the main paper, we continued with use of the ResNet models for a fair comparison to other methods.

\subsection{Results on additional datasets}

In the following pages we include figures with results across more datasets for all tasks. Results are in line with those in the main paper. One observation is that different methods show stronger performance on the dense prediction tasks depending on the amount of data and model.

\begin{figure*}[h]
\begin{center}
   \includegraphics[width=.95\linewidth]{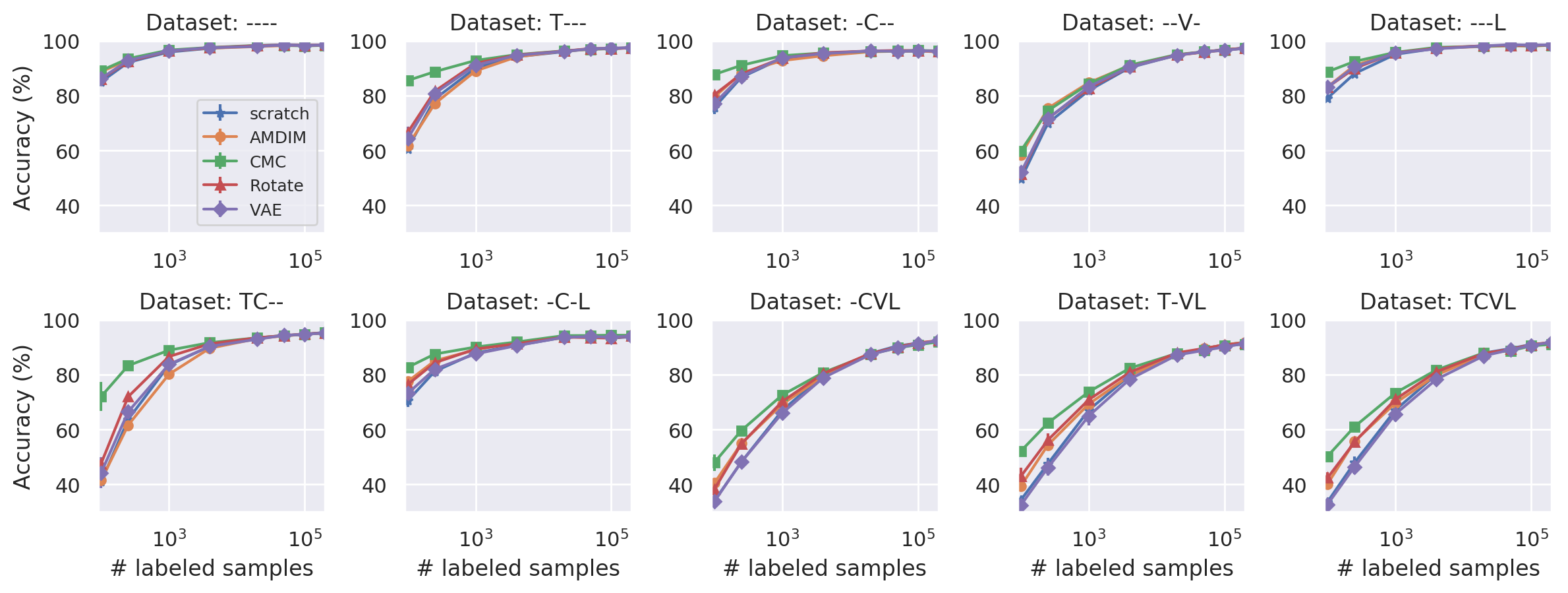}
\end{center}
\vspace{-.3cm}
   \caption{Object classification accuracy (ResNet9).}
\label{fig:all_cls_acc}
\end{figure*}

\begin{figure*}[h]
\begin{center}
   \includegraphics[width=.95\linewidth]{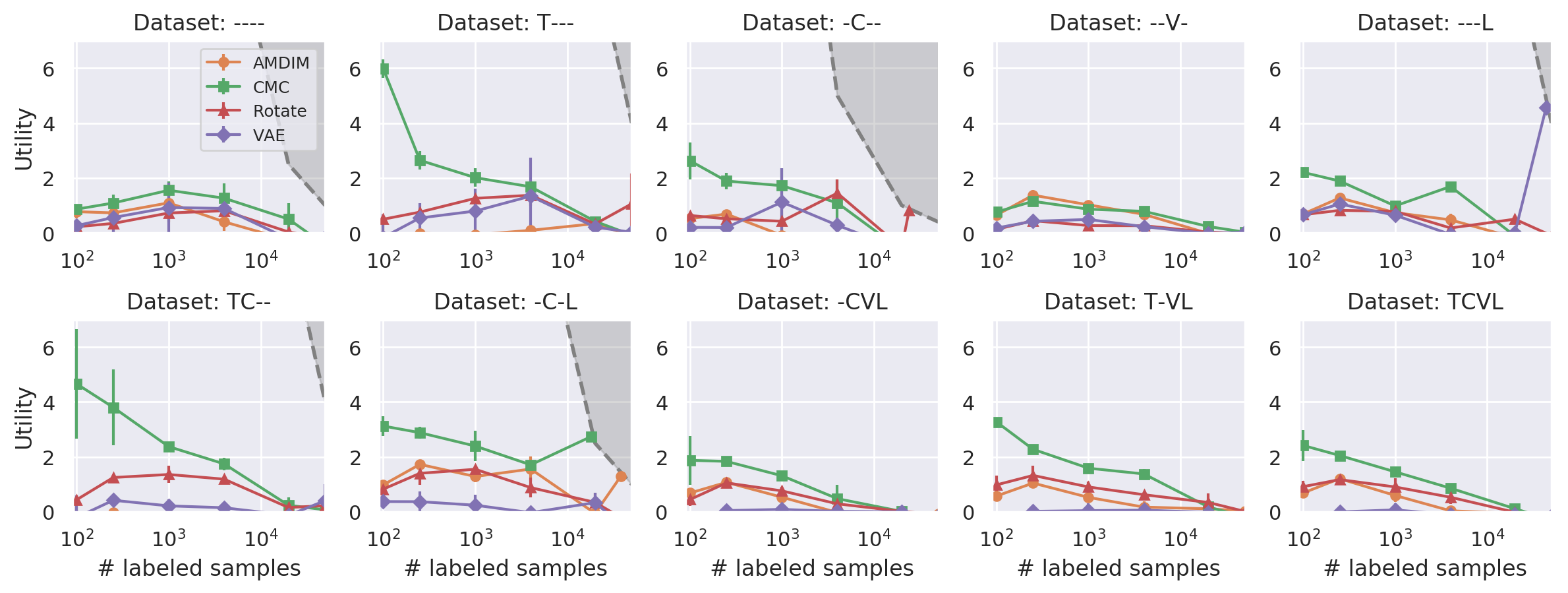}
\end{center}
\vspace{-.3cm}
   \caption{Object classification utility (ResNet9).}
\label{fig:all_cls_acc}
\end{figure*}

\begin{figure*}[h]
\begin{center}
   \includegraphics[width=.95\linewidth]{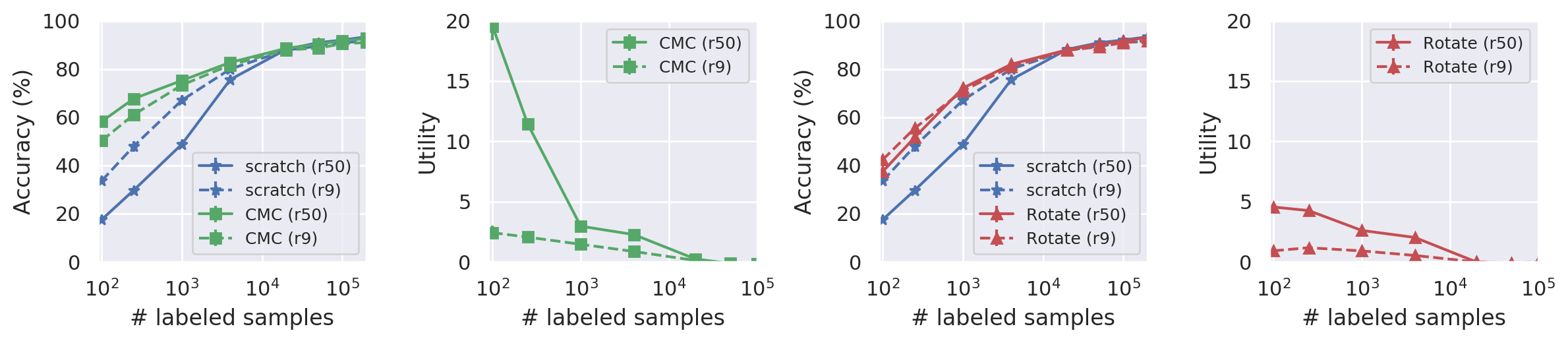}
\end{center}
\vspace{-.3cm}
   \caption{Object classification accuracy and utility (ResNet50).}
\label{fig:all_cls_acc}
\end{figure*}

\clearpage

\begin{figure*}[h]
\begin{center}
   \includegraphics[width=\linewidth]{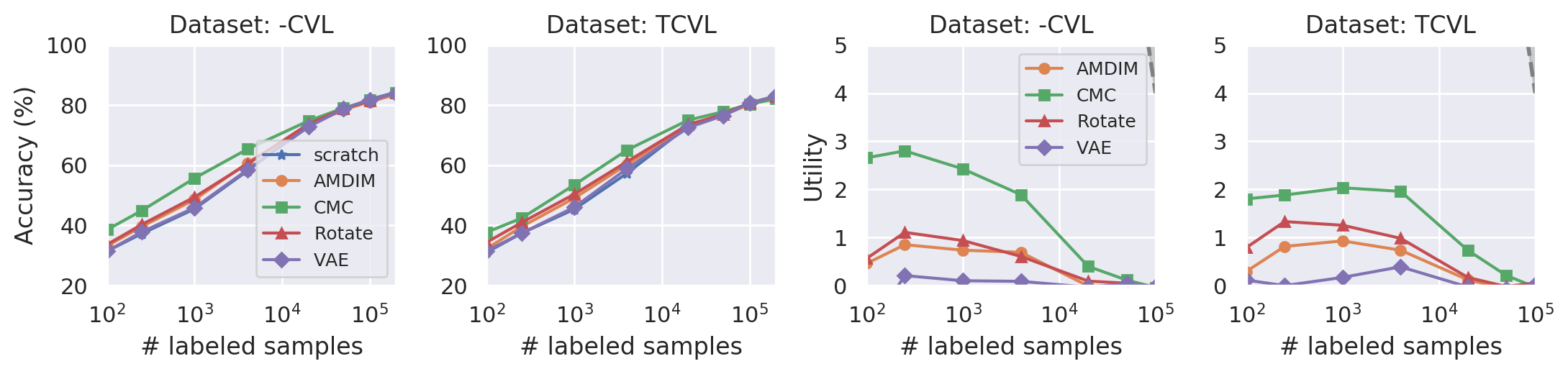}
\end{center}
\vspace{-.3cm}
   \caption{Object pose estimation results (ResNet9).}
\label{fig:all_cls_acc}
\end{figure*}

\begin{figure*}[h]
\begin{center}
   \includegraphics[width=\linewidth]{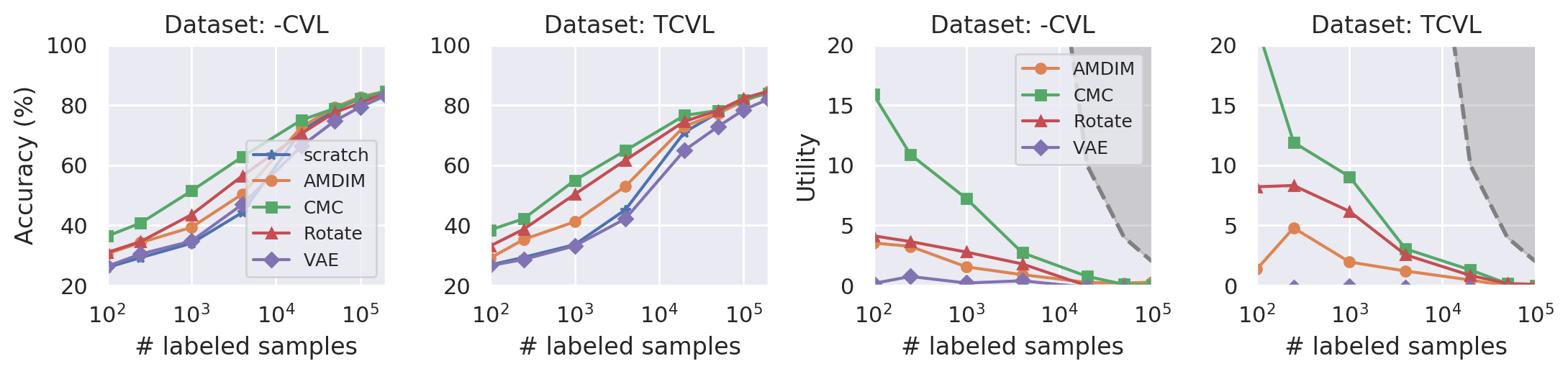}
\end{center}
\vspace{-.3cm}
   \caption{Object pose estimation results (ResNet50).}
\label{fig:all_cls_acc}
\end{figure*}

\begin{figure*}[h]
\begin{center}
   \includegraphics[width=\linewidth]{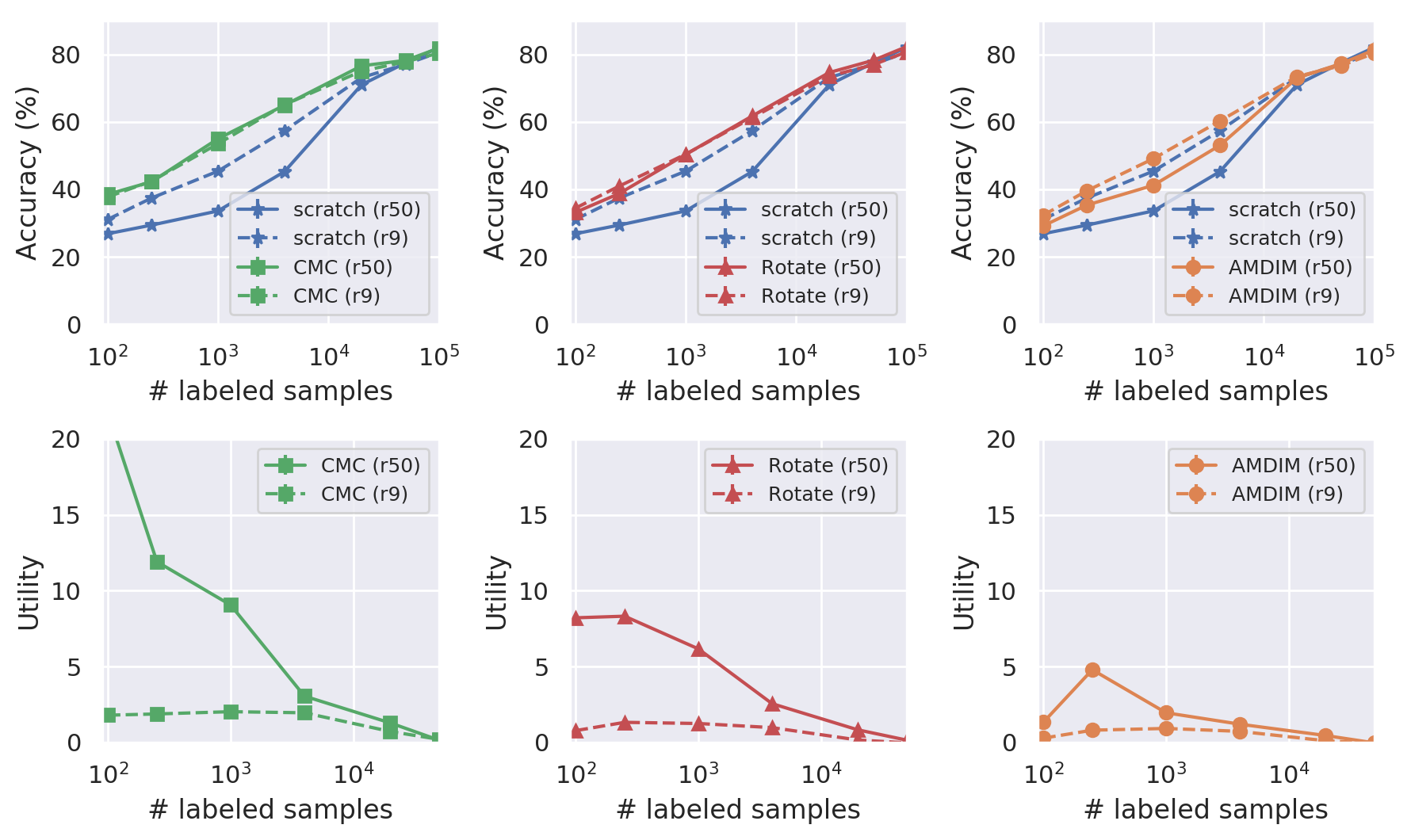}
\end{center}
\vspace{-.3cm}
   \caption{Direct comparison between ResNet9 and ResNet50 for object pose estimation.}
\label{fig:all_cls_acc}
\end{figure*}

\clearpage

\begin{figure*}[h]
\begin{center}
   \includegraphics[width=\linewidth]{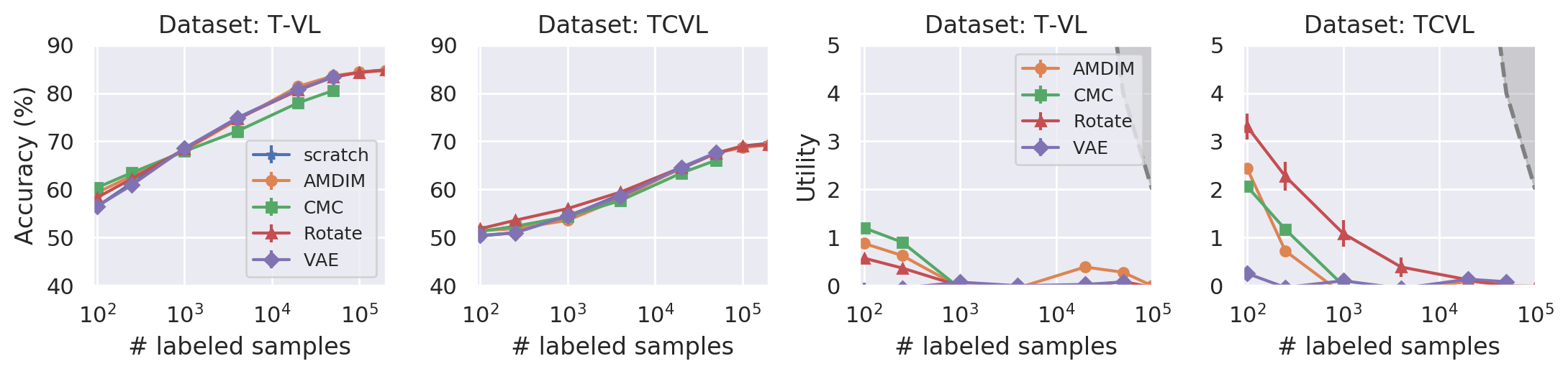}
\end{center}
\vspace{-.3cm}
   \caption{Semantic segmentation results (ResNet9).}
\label{fig:all_cls_acc}
\end{figure*}

\begin{figure*}[h]
\begin{center}
   \includegraphics[width=\linewidth]{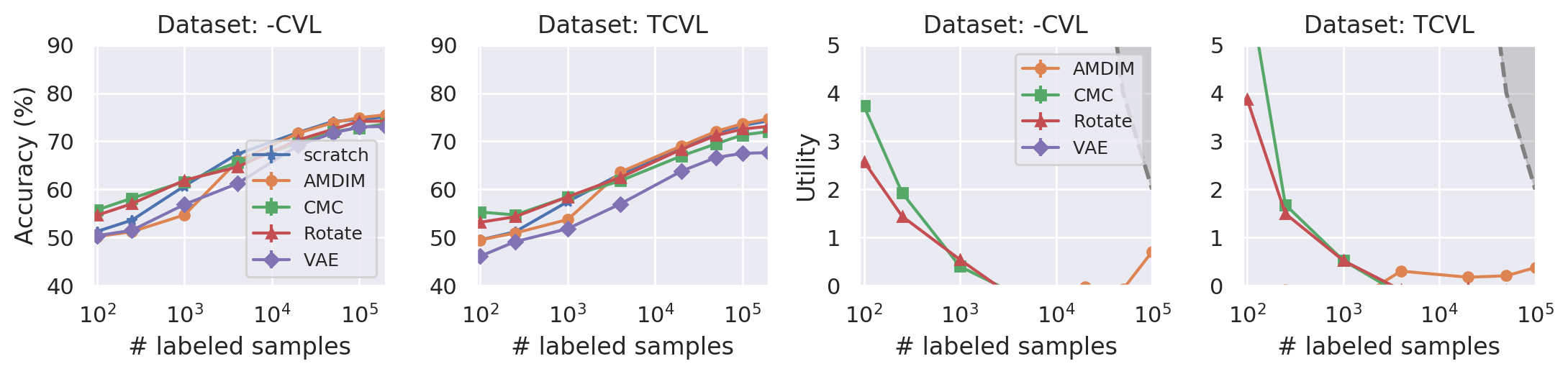}
\end{center}
\vspace{-.3cm}
   \caption{Semantic segmentation results (ResNet50).}
\label{fig:all_cls_acc}
\end{figure*}

\begin{figure*}[h]
\begin{center}
   \includegraphics[width=\linewidth]{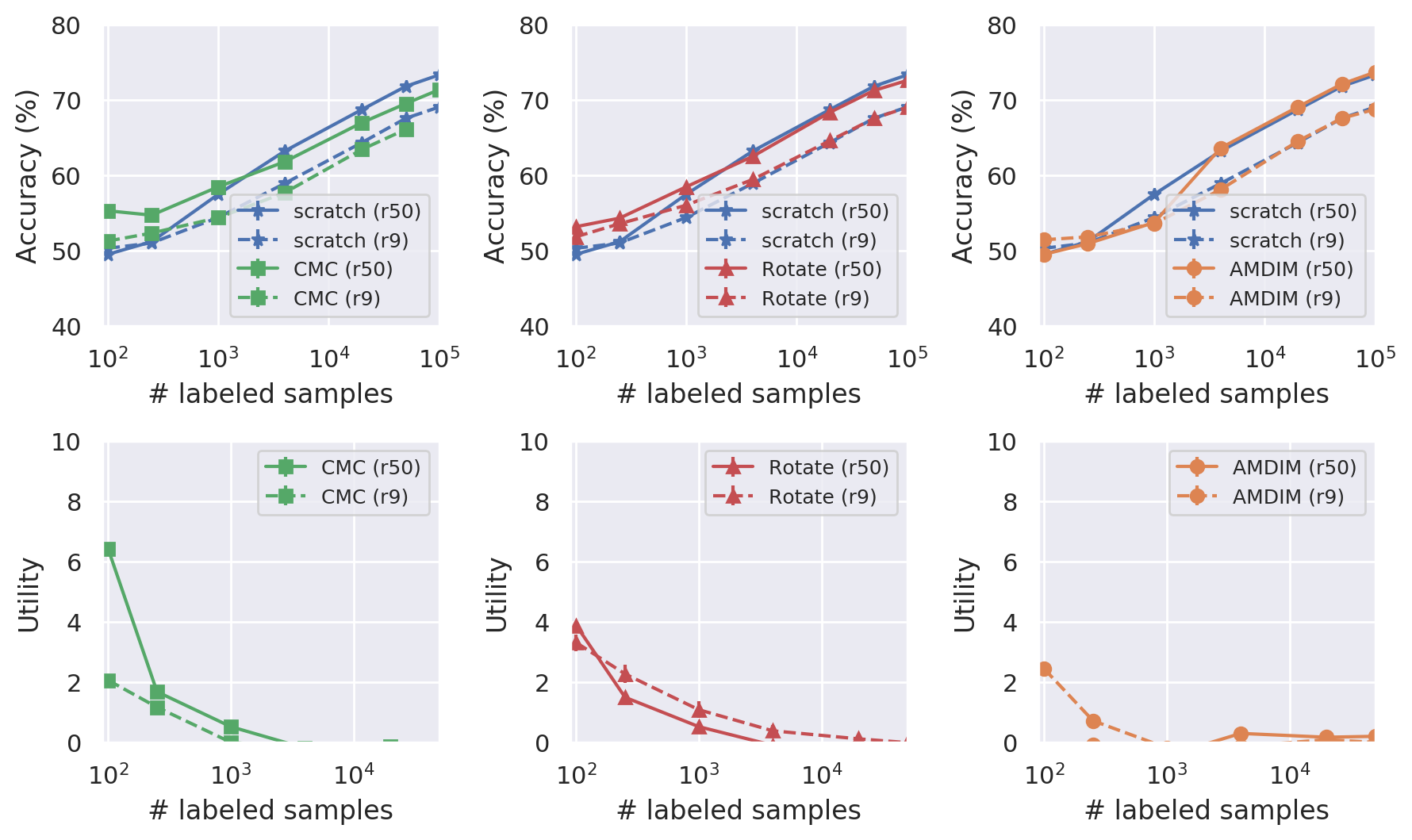}
\end{center}
\vspace{-.3cm}
   \caption{Direct comparison between ResNet9 and ResNet50 for semantic segmentation.}
\label{fig:seg_r9_r50}
\end{figure*}

\clearpage

\begin{figure*}[h]
\begin{center}
   \includegraphics[width=\linewidth]{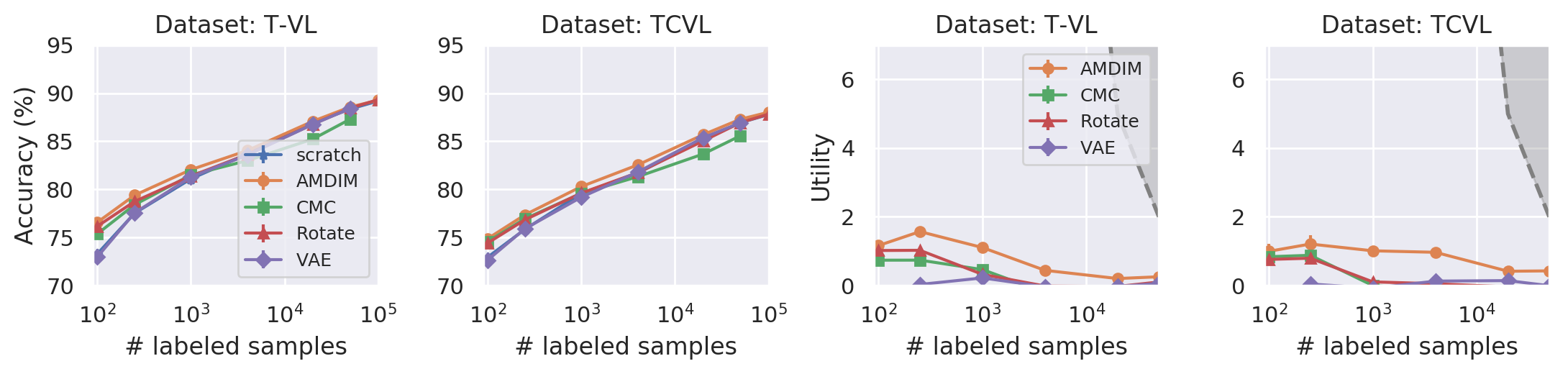}
\end{center}
\vspace{-.3cm}
   \caption{Depth estimation results (ResNet9).}
\label{fig:all_cls_acc}
\end{figure*}

\begin{figure*}[h]
\begin{center}
   \includegraphics[width=\linewidth]{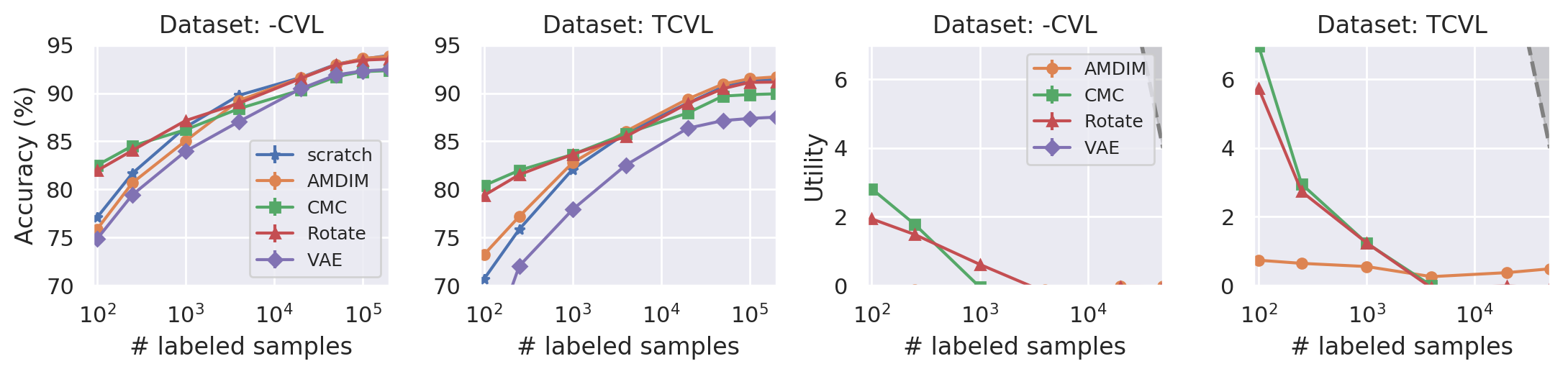}
\end{center}
\vspace{-.3cm}
   \caption{Depth estimation results (ResNet50).}
\label{fig:all_cls_acc}
\end{figure*}
\begin{figure*}
\begin{center}
   \includegraphics[width=\linewidth]{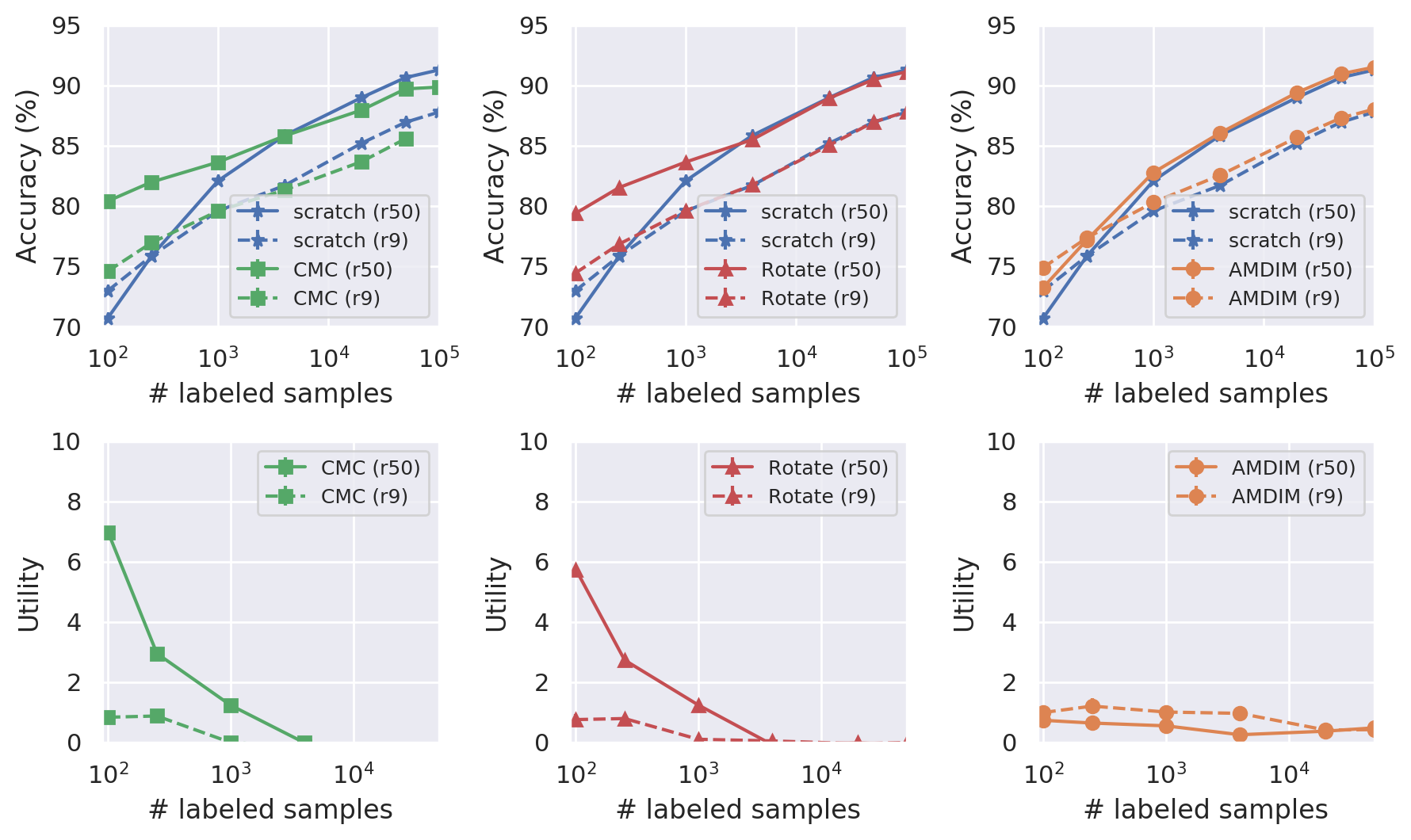}
\end{center}
\vspace{-.3cm}
   \caption{Direct comparison between ResNet9 and ResNet50 for depth estimation.}
\label{fig:depth_r9_r50}
\end{figure*}

\end{document}